\documentclass[conference]{IEEEtran}
\IEEEoverridecommandlockouts
\usepackage{amsmath,amssymb,amsfonts}
\usepackage{algorithm}
\usepackage{algpseudocode}
\usepackage{graphicx}
\usepackage{textcomp}
\usepackage{xcolor}
\usepackage{booktabs}
\usepackage[numbers]{natbib}

\newcommand{\Rnn}{\mathcal{R}_{\theta}}
\newcommand{\Rop}{\mathcal{R}}
\newcommand{\Fop}{\mathcal{F}}

\newcommand{\Jf}{\mathbf{J}_{\mathcal F}}
\newcommand{\m}{\mathbf{m}}
\newcommand{\mzero}{\mathbf{m}_0}
\newcommand{\dobs}{\mathbf{d}_{\mathrm{obs}}}

\def\BibTeX{{\rm B\kern-.05em{\sc i\kern-.025em b}\kern-.08em
    T\kern-.1667em\lower.7ex\hbox{E}\kern-.125emX}}

\begin{document}

\title{Full Waveform Inversion with a Learned Preconditioner}

\author{\IEEEauthorblockN{Jeongjin (Jayjay) Park}\\
jpark3141@gatech.edu}

\maketitle


\begin{abstract}
Full-waveform inversion (FWI) is a powerful approach for reconstructing subsurface velocity models, but its practical performance is limited by ill-conditioning, sensitivity to the update direction, and the high cost of curvature-based methods. In this project, we investigate whether a neural operator trained on reverse-time-migration (RTM) images can provide a useful learned preconditioner for FWI. We compare projected FWI against neural-preconditioned FWI using either a standard Fourier neural operator (FNO) or a Jacobian-informed FNO (JI-FNO). The neural preconditioner applies the normal operator of the learned RTM map to modify the true Devito FWI gradient. Our experiments show that the JI-FNO preconditioner achieves the lowest model error while maintaining data-residual decay comparable to baseline projected FWI. In contrast, the standard FNO preconditioner is unstable and gives worse model recovery, despite using the same neural-preconditioning framework. We also identify a failure mode of a nonlinear neural Gauss Newton-type inner solve: directly matching the neural RTM output to the FWI gradient creates an objective-space mismatch and does not yield a reliable  descent direction. These results suggest that forward RTM-image prediction alone is insufficient for learned preconditioning; accurate learned derivatives and objective-consistent use of the neural operator are critical for producing useful inversion updates.
\end{abstract}

\section{Introduction}
Seismic inversion estimates subsurface properties from recorded wavefield data. In full-waveform inversion (FWI) \citep{tarantola1984inverse,virieux2009overview}, the velocity model $\m$ is recovered by minimizing a  least-squares objective
\begin{equation}
    \Phi(\m)=\frac{1}{2}\|\Fop(\m)-\dobs\|_2^2,
    \label{eq:fwi-objective}
\end{equation}
where $\Fop$ denotes the wave-equation modeling operator and $\dobs$ denotes the observed seismic data. Its adjoint-state gradient is
\begin{equation}
    \nabla_\m \Phi(\m)=\Jf(\m)^\top\bigl(\Fop(\m)-\dobs\bigr),
    \label{eq:fwi-gradient}
\end{equation}
where $\Jf(\m)$ is the Born modeling operator. Although FWI uses the full waveform, the inverse problem is strongly ill-conditioned because seismic data have limited bandwidth, limited aperture, and nonuniform illumination. As a result, gradient-based updates can reduce the waveform residual while still producing inaccurate model updates, especially in deeper or poorly illuminated regions.

A common way to improve the update direction is to incorporate curvature information. The Gauss--Newton (GN) system for FWI is
\begin{equation}
    \bigl(\Jf^\top\Jf + \lambda I\bigr)\delta \m
    = -\Jf^\top\bigl(\Fop(\m)-\dobs\bigr),
    \label{eq:gn-system}
\end{equation}
where $\Jf^\top\Jf$ is the GN Hessian and $\lambda>0$ is a damping parameter \citep{pratt1998gauss,li2016modified}. However, applying this operator requires repeated Born and adjoint-Born wave-equation solves. This motivates cheaper approximate preconditioners that can improve the model-space update geometry without explicitly solving the full numerical GN problem.

This project studies a learned alternative motivated by prior work on Jacobian-informed neural operators for seismic image-domain inversion \citep{park2025ji}. In that work, a Fourier neural operator (FNO) \citep{li2020fourier,kovachki2023neural} was trained to approximate an RTM imaging map
\begin{equation}
    \Rop:(\m,\mzero)\mapsto \delta \m_{\mathrm{RTM}},
    \label{eq:rtm-map}
\end{equation}
where $\mzero$ is a smooth background model and $\delta \m_{\mathrm{RTM}}$ is the migrated image. The central observation was that forward RTM-image accuracy alone is not sufficient for inversion: if the learned Jacobian is inaccurate, gradient-based optimization through the neural operator can move in physically incorrect directions. Jacobian-informed FNO (JI-FNO) addresses this issue by supervising Jacobian-vector products along informative model-space directions.

The goal of this final project is to test whether the trained RTM neural operator can be reused downstream as a learned preconditioner for  FWI. Rather than replacing the physical FWI gradient, we compute the true Devito gradient and then modify the update direction using the Jacobian of either a standard FNO or a JI-FNO. The main question is whether the derivative-informed JI-FNO produces a more useful preconditioner than the standard FNO. Our results show that JI-FNO preconditioning gives the lowest model error, while the standard FNO preconditioner becomes unstable and degrades the recovered model. This supports the main hypothesis: when a neural operator is used as a preconditioner, derivative accuracy is critical for producing reliable curvature information.

\section{Background: Jacobian-Informed FNO}
\subsection{Learning an RTM imaging operator}
The prior JI-FNO model learns the map in Eq.~\eqref{eq:rtm-map}. The input is the pair $(\m,\mzero)$ and the output is the RTM image $\delta \m_{\mathrm{RTM}}$. A standard FNO is trained with the forward prediction loss
\begin{equation}
    \mathcal{L}_{\mathrm{FNO}}
    =
    \|\Rnn(\m,\mzero)-\delta \m_{\mathrm{RTM}}\|_2^2.
    \label{eq:fno-loss}
\end{equation}
Because the trained neural operator is differentiable, its Jacobian can be applied using automatic differentiation. This makes it attractive for inverse problems, where one wants to use the learned map not only for prediction but also for gradient-based updates.

However, inversion depends on the derivative of the operator. A neural operator may match RTM images in an $L_2$ sense while having an inaccurate Jacobian. This is problematic when the neural operator is used to compute gradients, preconditioners, or inverse updates.

\subsection{Derivative supervision along informative directions}
To improve derivative accuracy, JI-FNO adds Jacobian-vector-product supervision along globally informative directions. These directions are obtained from dominant eigendirections of an averaged GN/Fisher-like operator,
\begin{align}
    &\bar{\mathbf{H}}_{\mathrm{GN}}
    =\mathbb{E}_{(\m,\mzero)\sim\mu}
    \left[\mathbf{J}_{\mathcal F}(\m,\mzero)^\top
    \mathbf{J}_{\mathcal F}(\m,\mzero)\right], \\
    &\bar{\mathbf{H}}_{\mathrm{GN}}\bar{\mathbf{v}}_j
    =\lambda_j\bar{\mathbf{v}}_j .
    \label{eq:avg-hessian}
\end{align}
Let $\bar{\mathbf{V}}_r=[\bar{\mathbf{v}}_1,\ldots,\bar{\mathbf{v}}_r]$. The JI-FNO training objective is
\begin{align}
\mathcal{L}_{\mathrm{JI}}&(\m,\mzero)
=\|\Rnn(\m,\mzero)-\delta\m_{\mathrm{RTM}}\|_2^2 \nonumber\\
&\quad+\gamma\left\|
\frac{\partial \Rnn}{\partial \m}(\m,\mzero)\bar{\mathbf{V}}_r
-
\frac{\partial \Rop}{\partial \m}(\m,\mzero)\bar{\mathbf{V}}_r
\right\|_2^2 .
\label{eq:ji-loss}
\end{align}
Only a small fraction of derivative information is supervised, but prior image-domain experiments showed that this improves inversion metrics. Table~\ref{tab:prior-results} summarizes the prior result that motivated this project.

\begin{table}[t]
\centering
\caption{Prior image-domain inversion results motivating this project \cite{park2025ji}. The number of supervised directions is denoted by $r$.}
\label{tab:prior-results}
\begin{tabular}{lccc}
\toprule
Metric & FNO & JI-FNO & JI-FNO \\
\midrule
$r$ & -- & 8 & 64 \\
Model error (rel. $L_2$) & 0.29 & \textbf{0.21} & \textbf{0.21} \\
Data residual (rel. $L_2$) & 0.28 & 0.19 & \textbf{0.12} \\
Recovered SSIM & 0.50 & 0.68 & \textbf{0.76} \\
\bottomrule
\end{tabular}
\end{table}

\section{Theory: From RTM Jacobian to Learned Curvature}
The motivation for using an RTM neural operator as a preconditioner comes from the relation between RTM and the GN Hessian. Define an RTM imaging operator
\begin{equation}
    \mathrm{RTM}(\m)=\mathbf{J}_{\mathcal F}(\m)^\top \mathbf{r}(\m),
    \quad
    \mathbf{r}(\m)=\dobs-\Fop(\m).
\end{equation}
Differentiating gives
\begin{equation}
    \frac{\partial\,\mathrm{RTM}(\m)}{\partial\m}
    =
    \frac{\partial \mathbf{J}_{\mathcal F}(\m)^\top}{\partial\m}\mathbf{r}(\m)
    +
    \mathbf{J}_{\mathcal F}(\m)^\top
    \frac{\partial \mathbf{r}(\m)}{\partial\m} .
    \label{eq:rtm-diff-full}
\end{equation}
Since $\partial \mathbf{r}(\m)/\partial\m=-\mathbf{J}_{\mathcal F}(\m)$, and since the first term in Eq.~\eqref{eq:rtm-diff-full} is the residual-weighted second-derivative term neglected in the GN approximation, we obtain
\begin{equation}
    \frac{\partial\,\mathrm{RTM}(\m)}{\partial\m}
    \approx
    -\mathbf{J}_{\mathcal F}(\m)^\top \mathbf{J}_{\mathcal F}(\m)
    =-\mathbf{H}_{\mathrm{GN}}(\m).
    \label{eq:rtm-hessian}
\end{equation}
Thus, the Jacobian of an RTM map is approximately Hessian up to sign convention.

In the final implementation, however, the neural preconditioner uses the \emph{normal operator} of the learned RTM map,
\begin{equation}
    D\Rnn(\m,\mzero)^\top D\Rnn(\m,\mzero),
    \label{eq:neural-normal}
\end{equation}
not the RTM Jacobian alone. 


This distinction matters. If $D\Rnn$ already approximates a GN Hessian action, then Eq.~\eqref{eq:neural-normal} behaves spectrally like a squared Hessian. In principle, one could attempt to precondition directly with $D\Rnn$ or its symmetrized form. In practice, however, the learned Jacobian is only explicitly supervised along the $r$ informative directions used during training. Therefore, its accuracy is strongest on the probed subspace
\[
    \mathcal{V}_r=\mathrm{span}\{\mathbf{v}_1,\ldots,\mathbf{v}_r\},
\]
but it is not guaranteed to be accurate, symmetric, or positive definite on the full model space. We found that CG applied directly to the derivative-based operator was unstable, which is consistent with the learned Jacobian having unreliable curvature outside the supervised subspace. For this reason, we use the neural normal operator $D\Rnn^{\top}D\Rnn$, which is positive semidefinite by construction and more suitable for a CG-based inner solve. Thus, the resulting method is a learned image-domain normal-operator preconditioner that applies a strong spectral filtering to the FWI gradient using RTM sensitivity information.

\section{Method}
\subsection{Baseline projected FWI}
The baseline is projected FWI with gradients computed by Devito \citep{louboutin2017full}. Given observed data $\dobs$ and the nonlinear wave-equation modeling operator $\Fop$, we solve the TV-constrained waveform inversion problem
\begin{equation}
    \min_{\m}
    \frac{1}{2}
    \left\|
    \Fop(\m)-\dobs
    \right\|_2^2
    \quad
    \mathrm{s.t.}
    \:
    \m \in \mathcal{C}_{\mathrm{box}},
    \:
    \|\nabla \m\|_1 \leq \tau .
    \label{eq:tv-constrained-fwi}
\end{equation}
Here, $\mathcal{C}_{\mathrm{box}}$ constrains velocity to a physically admissible range and $\tau$ controls the allowable total variation. TV-constrained formulations are used to stabilize FWI by promoting piecewise-smooth models while preserving sharp interfaces \citep{esser2018total,peters2017constraints}. At each iteration, we compute the true adjoint-state gradient 
\begin{equation}
\mathbf{g}_k=\nabla\Phi(\m_k),   
\label{eq:true-gradient}
\end{equation}
take a descent step, project onto the box--TV feasible set, and restore the known water layer.

\subsection{Neural RTM normal-operator preconditioning}
Let $\Rnn$ denote either the standard FNO or the JI-FNO approximation to the RTM image map. At each FWI iteration, we first compute the true  gradient in Eq. \eqref{eq:true-gradient} using Devito. We then define the learned neural normal operator
\begin{equation}
    \mathbf{A}_{\theta,k}
    =
    D\Rnn(\m_k,\mzero)^\top
    D\Rnn(\m_k,\mzero).
\end{equation}
The preconditioned direction is obtained by approximately solving
\begin{equation}
    \left(\mathbf{A}_{\theta,k}+\lambda I\right)
    \mathbf{p}_{\theta,k}
    =
    \mathbf{g}_k,
    \label{eq:neural-normal-preconditioner}
\end{equation}
where $\lambda>0$ is a damping parameter. The matrix $\mathbf{A}_{\theta,k}$ is never formed explicitly. Instead, $D\Rnn q$ is computed by a Jacobian-vector product (JVP), and $D\Rnn^\top z$ is computed by a vector-Jacobian product (VJP), both through automatic differentiation.

After computing $\mathbf{p}_{\theta,k}$, the model is updated using the same projected FWI form,
\begin{equation}
    \m_{k+1}
    =
    \mathcal{R}_{\mathrm{water}}
    \left(
    \Pi_{\mathrm{box,TV}}
    \left(
    \m_k-\alpha \mathbf{p}_{\theta,k}
    \right)
    \right),
    \label{eq:neural-projected-update}
\end{equation}
where $\Pi_{\mathrm{box,TV}}$ denotes projection onto the box and TV constraints and $\mathcal{R}_{\mathrm{water}}$ restores the water layer.

\subsection{FNO versus JI-FNO preconditioners}
The final comparison uses two networks with the same FNO architecture. The standard FNO is trained only with Eq.~\eqref{eq:fno-loss}. The JI-FNO is trained with Eq.~\eqref{eq:ji-loss}, which directly supervises the learned Jacobian along informative directions. Since the preconditioner in Eq.~\eqref{eq:neural-normal-preconditioner} depends on the learned Jacobian, this comparison isolates whether derivative-informed training improves the quality of the preconditioned FWI update.

\subsection{L-BFGS variants}
We also test L-BFGS variants \cite{liu1989lbfgs} initialized by the neural preconditioned direction. These methods use the neural normal-operator solve as an initial inverse-curvature approximation and then update the inverse-Hessian estimate using limited-memory curvature pairs. In practice, the L-BFGS variants decreased the objective rapidly in early iterations but later became unstable or overfit the local objective. Therefore, we report them at the early stopping point of 16 iterations.

\begin{algorithm}[t]
\small
\caption{FWI with neural preconditioning}
\label{alg:neural-preconditioned-fwi}
\begin{algorithmic}[1]
\Require Initial model $\m_0$, data $\dobs$, background $\mzero$, trained $\Rnn$, stepsize $\alpha$, damping $\lambda$, TV budget $\tau$
\For{$k=0,1,\ldots,K-1$}
    \State Compute true FWI gradient $\mathbf{g}_k=\nabla\Phi(\m_k)$.
    \State Define $\mathbf{A}_{\theta,k}\mathbf{q}=D\Rnn(\m_k,\mzero)^\top D\Rnn(\m_k,\mzero)\mathbf{q}+\lambda \mathbf{q}$.
    \State Solve $\mathbf{A}_{\theta,k}\mathbf{p}_{\theta,k}=\mathbf{g}_k$.
    \If{using L-BFGS}
        \State Apply L-BFGS initialized with $\mathbf{A}_{\theta,k}^{-1}\mathbf{g}_k$
    \Else
        \State Set $\mathbf{p}_k \leftarrow \mathbf{p}_{\theta,k}$.
    \EndIf
    \State Projected update: $\tilde{\m}_{k+1}=\Pi_{\mathrm{box,TV}}(\m_k-\alpha\mathbf{p}_{\theta,k})$.
    \State Restore water layer: $\m_{k+1}=\mathcal{R}_{\mathrm{water}}(\tilde{\m}_{k+1})$.
\EndFor
\State \Return $\m_K$
\end{algorithmic}
\end{algorithm}

\section{Experimental Setup}
The experiments use a sliced BG Compass velocity model \cite{jones2012building} discretized on a $256\times256$ grid with grid spacing $\Delta x=\Delta z=12.5$ m. The velocity range is constrained to approximately $1.5$--$4.5$ km/s. Observed data are simulated with Devito using 32 shots and a 15 Hz Ricker wavelet. The initial model is obtained by Gaussian smoothing the true velocity model while preserving the water layer.

The final comparison includes five methods:
\begin{itemize}
    \item \textbf{FWI}: projected FWI with Devito gradients, box constraints, TV projection, and water-layer restoration.
    \item \textbf{FWI + JI-FNO precond}: projected FWI using the JI-FNO neural normal operator as a learned preconditioner.
    \item \textbf{FWI + JI-FNO + L-BFGS}: L-BFGS initialized by the JI-FNO preconditioned direction, reported at 16 iterations due to early overfitting.
    \item \textbf{FWI + FNO precond}: projected FWI using the standard FNO neural normal operator.
    \item \textbf{FWI + FNO + L-BFGS}: L-BFGS initialized by the standard FNO preconditioned direction, also reported at 16 iterations due to early overfitting.
\end{itemize}

We evaluate the inversion using the data misfit
\begin{equation}
    \mathrm{Data \: Error}_k
    =
    \left\|
    \mathcal{F}(\m_k)-\mathbf{d}_{\mathrm{obs}}
    \right\|_2^2 .
    \label{eq:dataerror}
\end{equation}
and the relative model error in percentage
\begin{equation}
    \mathrm{Model \: Error}_k
    = 100 \times 
    \frac{
    \|\m_k-\m_{\mathrm{true}}\|_2
    }{
    \|\m_{\mathrm{true}}\|_2
    \label{eq:modelerror}
    }.
\end{equation}
These two metrics need not decrease at the same rate because FWI is nonlinear and ill-conditioned.

Projected FWI required approximately 35 minutes in this setup, while neural-preconditioned FWI required approximately 1 hour and 35 minutes due to the inner neural Krylov solves. Thus, the learned preconditioner is more expensive than plain FWI, but it avoids wave-equation Hessian-vector products and is intended as a cheaper alternative to full numerical curvature methods.

\section{Results and Discussion}
\begin{figure*}[t]
    \centering
    \includegraphics[width=0.95\textwidth]{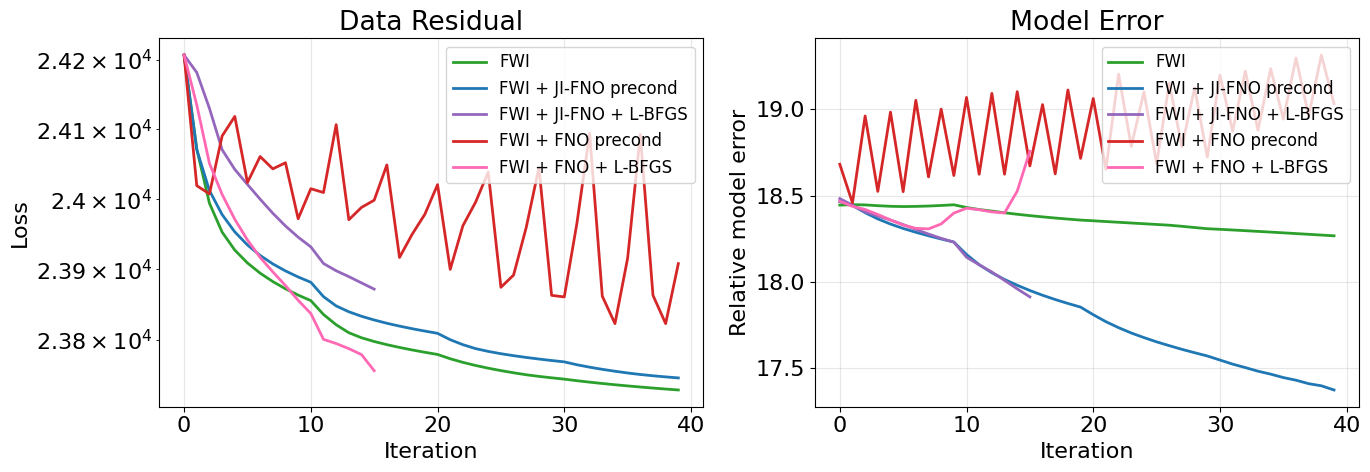}
    \caption{Data residual and model-error curves for projected FWI and neural-preconditioned variants. The metrics are defined in Eq.~\eqref{eq:modelerror} and Eq.~\eqref{eq:dataerror}. Baseline FWI achieves the lowest final data residual, but JI-FNO preconditioning gives the lowest model error. The standard FNO preconditioner is unstable, showing oscillatory data residuals and increasing model error. L-BFGS variants are shown only up to early stopping because they overfit after several iterations.}
    \label{fig:metrics}
\end{figure*}

\subsection{Quantitative comparison}
Table~\ref{tab:final-results} summarizes the final inversion metrics. Baseline FWI obtains the lowest data misfit, but the JI-FNO-preconditioned method obtains the lowest model error. Specifically, the model error decreases from 18.2669 for baseline FWI to 17.3743 for FWI with JI-FNO preconditioning, while the final data residual remains comparable. This indicates that reducing the waveform residual fastest does not necessarily produce the most accurate velocity model.

\begin{table}[t]
\centering
\caption{Final inversion metrics. Groundtruth velocity model and the recovered models are shown in Figure \ref{fig:models}.}
\label{tab:final-results}
\begin{tabular}{lccc}
\toprule
Method & Iter. & Final data misfit & Final model error \\
\midrule
FWI & 40 & $\mathbf{2.3729\times 10^4}$ & $18.2669$ \\
FWI + JI-FNO precond & 40 & $2.3746\times 10^4$ & $\mathbf{17.3743}$ \\
FWI + JI-FNO + L-BFGS & 16 & $2.3872\times 10^4$ & $17.9131$ \\
FWI + FNO precond & 40 & $2.3908\times 10^4$ & $19.0320$ \\
FWI + FNO + L-BFGS & 16 & $2.3756\times 10^4$ & $18.7563$ \\
\bottomrule
\end{tabular}
\end{table}

\subsection{Data residual versus model error}
Figure~\ref{fig:metrics} shows that data residual and model error can behave differently. Baseline FWI steadily decreases the waveform residual, but its model error improves less than the JI-FNO-preconditioned result. In contrast, JI-FNO preconditioning produces a slightly slower residual decay but substantially better model recovery. This behavior is consistent with a learned model-space preconditioner: it changes the geometry of the update rather than simply following the steepest waveform-data descent direction.

The standard FNO preconditioner performs substantially worse. Its data residual oscillates and its final model error increases to 19.0320. This suggests that an FNO trained only for forward RTM-image prediction does not provide a reliable Jacobian for preconditioning. 


\begin{figure*}[t]
    \centering
    \includegraphics[width=0.95\textwidth]{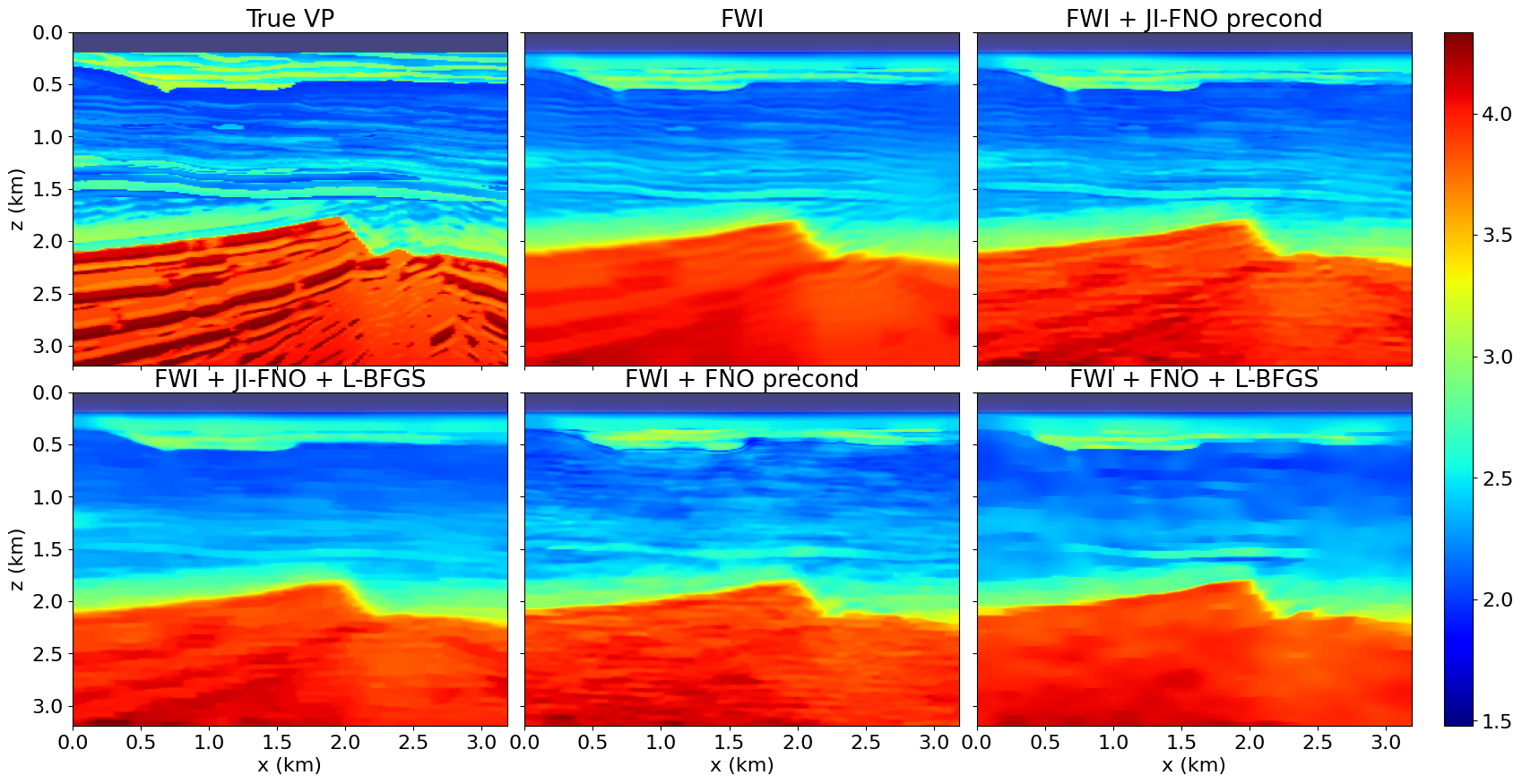}
    \caption{Recovered velocity models for projected FWI and neural-preconditioned projected FWI methods. The JI-FNO preconditioned result better preserves the large-scale velocity structure and reduces model error compared with baseline FWI. In contrast, the standard FNO preconditioner introduces artifacts and degrades the recovered model, indicating that forward RTM-image training alone is insufficient for reliable preconditioning.}
    \label{fig:models}
\end{figure*}



\subsection{Qualitative model comparison}

Figure~\ref{fig:models} compares the recovered velocity models for the baseline and neural-preconditioned methods. The JI-FNO-preconditioned result recovers the deep high-velocity structure more clearly than baseline FWI and the standard FNO-preconditioned result. In particular, the mid-depth region, between 0.8 km and 1.8 km, is better preserved by the JI-FNO preconditioner: the recovered layers remain more continuous and closer to the reference model (True VP), while still maintaining a level of detail comparable to projected FWI.

\begin{figure*}[t]
    \centering
    \includegraphics[width=0.95\textwidth]{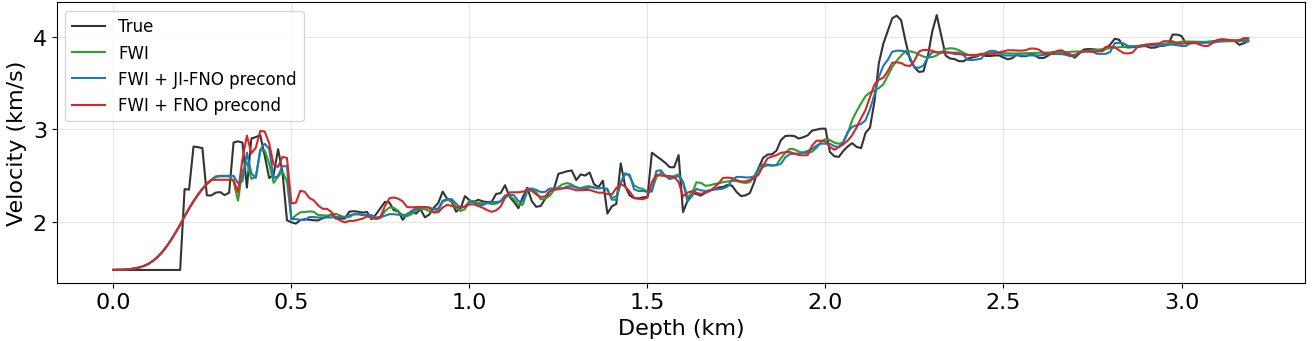}
    \caption{Vertical velocity trace comparison at a lateral location of 2.5 km. 
    The JI-FNO-preconditioned result follows the reference velocity trace more closely over several depth intervals, indicating improved recovery of the layered structure. 
    In contrast, the standard FNO-preconditioned result shows deviations, particularly around 0.5 km depth, and overly smooths several velocity contrasts.
    }
    \label{fig:trace}
\end{figure*}

The trace comparison in Figure~\ref{fig:trace} further supports this observation. The JI-FNO-preconditioned trace follows the reference velocity (FWI) more closely over several depth intervals, especially in the middle and deeper parts of the model. In contrast, the standard FNO preconditioner introduces stronger artifacts and excessive smoothing. This smoothing blurs layer boundaries and weakens the deeper velocity contrasts (Figure~\ref{fig:models}), which explains its larger final model error. These qualitative results are consistent with the quantitative metrics: Jacobian-informed training produces a more useful neural preconditioner, whereas forward-only FNO training does not provide sufficiently reliable derivative information for stable model recovery.

\subsection{Behavior of the L-BFGS variants}
The L-BFGS variants improve model error during early iterations but become less reliable afterward. The JI-FNO + L-BFGS variant achieves a model error of 17.9131 at 16 iterations, which improves over baseline FWI but does not outperform direct JI-FNO preconditioning. The FNO + L-BFGS variant gives a lower data residual than direct FNO preconditioning, but its model error remains worse than both FWI and JI-FNO preconditioning. This suggests that L-BFGS can accelerate local objective reduction, but it is sensitive to the quality of the initial preconditioned direction.



\subsection{Interpretation}

The results indicate that the neural normal operator acts less like an exact  Gauss--Newton Hessian and more like a learned image-domain regularizer. Since $D\Rnn^\top D\Rnn$ applies additional spectral filtering to the RTM sensitivity, it can suppress harmful update components and improve the model-space trajectory, even when it does not reduce the data misfit faster than baseline FWI.

The contrast between FNO and JI-FNO is central: both methods use the same preconditioning framework, but only JI-FNO is trained to match derivative information. The improved behavior of JI-FNO therefore supports the conclusion that derivative accuracy is essential when neural operators are used to define optimization updates.

\subsection{Failure mode of the nonlinear neural inner solve}

We also considered a nonlinear neural inner-solve strategy that is conceptually different from the neural normal-operator preconditioner used in the final experiments. In this variant, the current FWI gradient is treated as an image-domain target, and we attempt to find a model $\mathbf{z}$ satisfying
\begin{equation}
    \Rnn(\mathbf{z},\mzero)
    \approx
    \mathbf{g}_{\mathrm{FWI}}(\m_k),
    \label{eq:failed-nonlinear-inner}
\end{equation}
then use $\mathbf{z}-\m_k$ as the model update. This method did not work reliably because Eq.~\eqref{eq:failed-nonlinear-inner} solves a nonlinear RTM-image matching problem, whereas the outer FWI objective is defined in waveform-data space.

To see the mismatch, recall that the  Gauss--Newton update satisfies
\begin{equation}
    \left(
    \mathbf{J}_{F,k}^{\top}\mathbf{J}_{F,k}
    +
    \lambda I
    \right)
    \mathbf{p}^{\mathrm{GN}}_k
    =
    -
    \mathbf{J}_{F,k}^{\top}
    \left(
    \Fop(\m_k)-\dobs
    \right),
    \label{eq:true-gn-normal}
\end{equation}
where $\mathbf{J}_{F,k}=D\Fop(\m_k)$ is the Born modeling operator. Thus, the true Gauss--Newton direction is constructed from the waveform-data residual and the waveform-data Jacobian.

By contrast, the nonlinear neural inner solve defines an RTM-image matching problem. Let
\[
    \mathbf{J}_{\theta,k}
    =
    D_{\mathbf{z}}\Rnn(\mathbf{z},\mzero)
\]
denote the Jacobian of the neural RTM map at the current inner iterate. The local Gauss--Newton normal equation associated with the inner problem is
\begin{equation}
    \mathbf{J}_{\theta,k}^{\top}
    \mathbf{J}_{\theta,k}
    \Delta \mathbf{z}_k
    =
    -
    \mathbf{J}_{\theta,k}^{\top}
    \left(
    \Rnn(\mathbf{z},\mzero)
    -
    \mathbf{g}_{\mathrm{FWI}}(\m_k)
    \right).
    \label{eq:failed-inner-normal}
\end{equation}
This resembles a Gauss--Newton system, but it is the Gauss--Newton system for an RTM-image matching problem, not for  FWI. The residual in Eq.~\eqref{eq:failed-inner-normal} is
$\Rnn(\mathbf{z},\mzero)-\mathbf{g}_{\mathrm{FWI}}(\m_k)$,
whereas the residual in Eq.~\eqref{eq:true-gn-normal} is
$\Fop(\m_k)-\dobs$. Therefore, the update $\mathbf{z}-\m_k$ need not be a descent direction for the FWI objective.

\section{Conclusion}
This project investigated whether a neural operator trained on RTM images can be used as a learned preconditioner for  FWI. The final experiments compare projected FWI with neural-preconditioned FWI using either a standard FNO or a Jacobian-informed FNO. The results show that JI-FNO preconditioning gives the lowest model error while maintaining data-residual decay comparable to baseline FWI. In contrast, the standard FNO preconditioner is unstable and leads to worse model recovery.

These results support the main hypothesis that accurate learned derivatives are necessary when neural operators are used inside inverse-problem optimization. A standard FNO may predict RTM images well, but its Jacobian is not constrained to represent the physical sensitivity of the imaging map. Because the preconditioner depends directly on the neural Jacobian, derivative error can lead to poor update directions. The JI-FNO mitigates this issue by explicitly supervising Jacobian-vector products along informative directions.

The learned preconditioner should not be interpreted as an exact  Gauss--Newton method. It uses the normal operator of the learned RTM map, which acts as an image-domain model-space preconditioner. Future work should investigate line-search strategies, better scaling of the neural preconditioned direction, and image-domain or LSRTM formulations where the learned RTM geometry is more directly aligned with the optimization objective.

\section{Acknowledgment}
This report was drafted from the course presentation \emph{Full Waveform Inversion with Learned Preconditioner} and the prior abstract/report \emph{Neural Operators with Accurate Jacobian for High-Fidelity Image-Domain Seismic Inversion}. The author reviewed and remains responsible for the final content.

\bibliographystyle{plainnat}
\bibliography{references_overleaf}

\end{document}